\documentclass{article} % For LaTeX2e
\usepackage{iclr2020_conference,times}

\usepackage{amsmath,amsfonts,bm}

\def\eqref#1{equation~\ref{#1}}
\def\1{\bm{1}}

\DeclareMathAlphabet{\mathsfit}{\encodingdefault}{\sfdefault}{m}{sl}
\SetMathAlphabet{\mathsfit}{bold}{\encodingdefault}{\sfdefault}{bx}{n}

\usepackage{enumitem}
\usepackage{hyperref}
\usepackage{url}
\usepackage{xspace}
\usepackage{graphicx}
\usepackage{subfigure}
\usepackage{xcolor}
\usepackage{multirow}
\usepackage{comment}
\usepackage{float}
\makeatletter
\renewcommand{\thesubfigure}{\alph{subfigure}}% (a) -> a
\renewcommand{\@thesubfigure}{(\thesubfigure)\hskip\subfiglabelskip}% a -> a)
\makeatother
\definecolor{myred}{rgb}{0.8,0,0}
\definecolor{myblue}{rgb}{0.14,0.38,0.52}

\newcommand{\deepset}{Deep Sets\xspace}
\usepackage{wrapfig}
\newcommand{\bi}[1]{\textbf{\textit{#1}}}
\usepackage{multicol}
\newcommand{\imagine}{\textsc{imagine}\xspace}
\newcommand{\balpha}{\boldsymbol{\alpha}}
\newcommand{\bbeta}{\boldsymbol{\beta}}
\newcommand{\bgamma}{\boldsymbol{\gamma}}
\newcommand{\bDelta}{\boldsymbol{\Delta}}

\makeatletter
\let\@fnsymbol\@arabic
\makeatother

\title{Deep Sets for Generalization in RL}

\author{Tristan Karch\thanks{for equal contribution} \textit{} \footnotemark[2]\\
\And
Cedric Colas\footnotemark[1] \textit{ }\thanks{Flowers Team, INRIA, France. Correspondence to: Tristan Karch $<$tristan.karch@inria.fr$>$}\\
\And
Laetitia Teodorescu \footnotemark[2]  \\
\And
Cl\'ement Moulin-Frier \footnotemark[2]\\
\AND
Pierre-Yves Oudeyer \footnotemark[2]\\
}

\iclrfinalcopy % Uncomment for camera-ready version, but NOT for submission.
\begin{document}

\maketitle

\begin{abstract}
This paper investigates the idea of encoding object-centered representations in the design of the reward function and policy architectures of a language-guided reinforcement learning agent. This is done using a combination of object-wise permutation invariant networks inspired from Deep Sets and gated-attention mechanisms. In a 2D procedurally-generated world where agents targeting goals in natural language navigate and interact with objects, we show that these architectures demonstrate strong generalization capacities to out-of-distribution goals. We study the generalization to varying numbers of objects at test time and further extend the object-centered architectures to goals involving relational reasoning.
\end{abstract}

% % % % % % % % % % % % % % % % % % % % % % % % % % % % % % % % % % % % % % % % % % % % % % % % % % % % % %
\section{Introduction}
% % % % % % % % % % % % % % % % % % % % % % % % % % % % % % % % % % % % % % % % % % % % % % % % % % % % % %

Reinforcement Learning (RL) has begun moving away from simple control tasks to more complex multimodal environments, involving compositional dynamics and language. To successfully navigate and represent these environments, agents can leverage factorized representations of the world as a collection of constitutive elements. Assuming objects share properties \citep{green2017object}, agents may transfer knowledge or skills about one object to others. Just like convolutional networks are tailored for images, relational inductive biases \citep{battaglia2018relational} can be used to improve reasoning about relations between objects (e.g. in the CLEVR task, \cite{Johnson_2017}). One example could be to restrict operations to inputs related to object pairs.

% Furthermore, problems involving relational reasoning (reason about relations between objects like in the CLEVR task, \cite{Johnson_2017}) become easier by operating on pairs of objects, providing the relational inductive biases \citep{battaglia2018relational} required by many real-world settings.

A companion paper described a setting where an agent that sets its own goals has to learn to interact with a set of objects while receiving descriptive feedbacks in natural language (NL) \citep{imagine}. This work introduced reward function and policy architectures inspired by \deepset \citep{deepset} which operate on unordered sets of object-specific features, as opposed to the traditional concatenation of the features of all objects. In this paper, we aim to detail that contribution by studying the benefits brought by such architectures. We also propose to extend them to consider pairs of objects, which provides inductive biases for language-conditioned relational reasoning.
%These architectures were shown to result in high sample efficiency and strong generalizations to out-of-distribution goals.

In these architectures, the final decision (e.g. reward, action) integrates sub-decisions taken at the object-level. Every object-level decision takes into account relationships between the body --a special kind of object-- and either one or a pair of external objects. This addresses a core issue of language understanding \citep{kaschak2000constructing,bergen2015embodiment}, by grounding the meaning of sentences in terms of affordant relations between one's body and external objects. 

In related work, \cite{santoroRBMPBL17} introduced Relational Networks, language-conditioned relational architectures used to solve the supervised CLEVR task. \cite{zambaldi2018relational} introduced relational RL by using a Transformer layer \citep{vaswani2017attention} to operate on object pairs, but did not use language. Our architectures also draw inspiration from gated attention mechanisms \citep{chaplot2017gatedattention}. Although other works also propose to train a reward function in parallel of the policy, they do so using domain knowledge (expert dataset in \cite{bahdanau2018learning}, environment dynamics in \cite{fu2018from}) and do not leverage object-centered representations.

\paragraph{Contributions -}In this paper, we study the comparative advantage of using architectures based on factorized object representations for learning policies and reward functions in a language-conditioned RL setting. We 1) prove that our proposed architectures perform best in this setting compared to non-factorized baselines, 2) study their capacity to allow generalization to out-of-distribution goals and generalization to additional objects in the scene at test time, and 3) show that this architecture can be extended to deal with goals related to object pairs.

% % % % % % % % % % % % % % % % % % % % % % % % % % % % % % % % % % % % % % % % % % % % % % % % % % % % % %
\section{Problem Definition}
% % % % % % % % % % % % % % % % % % % % % % % % % % % % % % % % % % % % % % % % % % % % % % % % % % % % % %
A learning agent explores a procedurally generated 2D world containing objects of various types and colors. Evolving in an environment where objects share common properties, the agent can transfer knowledge and skills between objects, which enables \textit{systematic generalization} (e.g. \textit{grasp green tree + grasp red cat $\to$ grasp red tree}). The agent can navigate in the 2D plane, grasp objects and grow some of them (animals and plants). A simulated social partner (SP) provides NL labels when the agent performs interactions that SP considers interesting (e.g. \textit{grasp green cactus}). Descriptions are turned into targetable goals by the agent and used to train an internal reward function. Achievable goals $\mathcal{G}^\text{A}$ are generated according to the following grammar: 
\vspace{-0.3cm}
\begin{multicols}{2}
\begin{enumerate}[leftmargin=0.6cm, noitemsep]
    \item Go: \textit{(e.g. go bottom left)} %\vspace{-0.2cm}
    \begin{itemize} [leftmargin=0.2cm, noitemsep]
            \item \textit{go} + \bi{zone}
            \end{itemize}
    \item Grasp: \textit{(e.g. grasp red cat)} %\vspace{-0.2cm}
        \begin{itemize}[leftmargin=0.2cm, noitemsep]
            \item \textit{grasp} + \bi{color} $\cup$ \{\textit{any}\}  + \textbf{\textit{object type $\cup$ object category}}
            \item \textit{grasp} + \textit{any} + \bi{color} + \textit{thing}
        \end{itemize}
    \item Grow: \textit{(e.g. grow green animal)} %\vspace{-0.2cm}
        \begin{itemize}[leftmargin=0.2cm, noitemsep]
            \item \textit{grow} + \bi{color} $\cup$ \{\textit{any}\} + \bi{living thing} $\cup$ \{\textit{living\_thing, animal, plant}\}
            \item \textit{grow} + \textit{any} + \bi{color} + \textit{thing}
    \end{itemize}
\end{enumerate} 
\end{multicols}
\vspace{-0.4cm}
\textbf{Bold} and \{ \} represent sets of words while \textit{italics} represents specific words, see detailed grammar in Section~\ref{sec:grammar}. In total, there are $256$ achievable goals, corresponding to an infinite number of scenes. These are split into a training set of goals $\mathcal{G}^\text{train}$ from which SP can provide feedbacks, and a testing set of goals $\mathcal{G}^\text{test}$ held out to test the generalization abilities of the agent. Although testing sentences are generated following the same composition rules as training sentences, they extend beyond the training distribution (\textit{out-of-distribution generalization}). The agent can show two types of generalizations: from the reward function (\textit{it knows when the goal is achieved}) and from the policy (\textit{it knows how to achieve it}). Full details about the setup, architectures and training schedules are reported from \cite{imagine} in the Appendices.

\paragraph{Evaluation -}Regularly, the agent is tested offline on goals from $\mathcal{G}^\text{train}$ (training performance) and goals from $\mathcal{G}^\text{test}$ (testing performance). We test both the average success rate $(\overline{SR})$ of the policy and the average $F_1$-score of the reward function over each set of goals. A goal's $SR$ is computed over $30$ evaluations. The $F_1$-score is computed from a held out set of trajectories (see Section~\ref{sec:supervised-dataset}). Note that \textit{training performance} refers to the performance on $\mathcal{G}^\text{train}$, but still measures \textit{state generalization} as scenes are generated procedurally. In all experiments, we provide the mean and standard deviation over $10$ seeds, and report statistical significance using a two-tail Welch's t-test at level $\alpha=0.05$ as advised in \citet{colas2019hitchhiker}.

% % % % % % % % % % % % % % % % % % % % % % % % % % % % % % % % % % % % % % % % % % % % % % % % % % % % % %
\section{Deep Sets for RL}
% % % % % % % % % % % % % % % % % % % % % % % % % % % % % % % % % % % % % % % % % % % % % % % % % % % % % %

\begin{figure}[!h]
      \centering
      \includegraphics[width=0.98\textwidth]{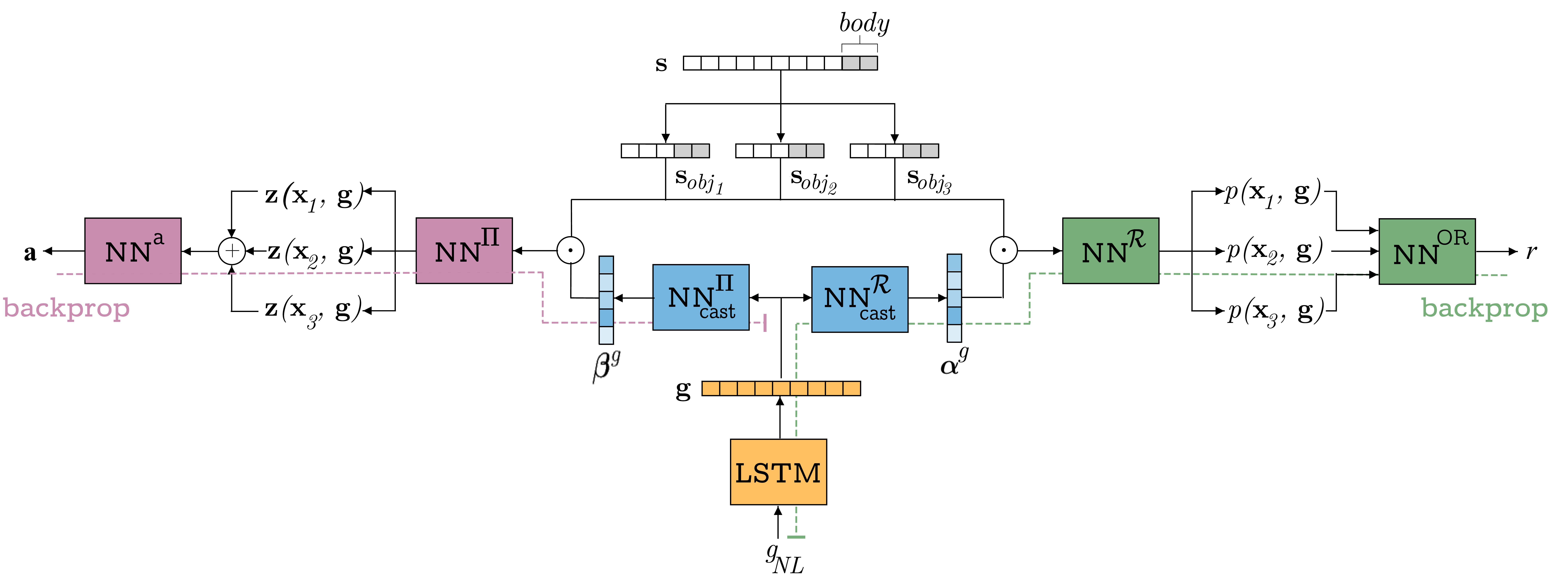}
      \caption{\textbf{Modular architectures with attention.} Left: policy. Right: reward function.}
      \label{fig:archis}
\end{figure}

The agent learns in parallel a language model, an internal goal-conditioned reward function and a multi-goal policy. The language model $(\mathcal{LM})$ embeds NL goals $(\mathcal{LM}(g_\text{NL}):\mathcal{G}^\text{A} \to \mathbb{R}^{100})$ using an LSTM \citep{hochreiter1997lstm} trained jointly with the reward function via backpropagation (yellow in Fig.~\ref{fig:archis}). The reward function, policy and critic become language-conditioned functions when coupled with $\mathcal{LM}$, which acts like a goal translator. The agent keeps tracks of goals discovered through its exploration and SP's feedbacks $\mathcal{G}_d$. It samples targets uniformly from $\mathcal{G}_d$. 

% The agent keeps tracks of goals discovered through its exploration and SP's feedbacks $\mathcal{G}_d$, and samples targets uniformly from $\mathcal{G}_d$. 

\paragraph{Deep Sets -}The reward function, policy and critic leverage modular architectures inspired by \deepset \citep{deepset} combined with gated attention mechanisms \citep{chaplot2017gatedattention}. \deepset is a family of neural architectures implementing set functions (input permutation invariance). Each input is mapped separately to some (usually high-dimensional \citep{wagstaff2019limitations}) latent space using a shared network. These latent representations are then passed through a permutation-invariant function (e.g. mean, sum) to ensure the permutation-invariance of the whole function.

\paragraph{Modular-attention architecture for the reward function -} Learning a goal-conditioned reward function $(\mathcal{R})$ is framed as binary classification. The reward function maps a state $\mathbf{s}$ and a goal embedding $\mathbf{g}=\mathcal{LM}(g_\text{NL})$ to a binary reward: $\mathcal{R}(\mathbf{s}, \mathbf{g}): \mathcal{S} \times \mathbb{R}^{100} \to \{ 0, 1\}$ (right in Fig.~\ref{fig:archis}). The reward function is constructed such that object-specific rewards are computed independently for each of the $N$ objects before being integrated into a global reward through a logical OR function approximated by a differentiable network which ensures object-wise permutation invariance: \textit{if any object verifies the goal, then the whole scene verifies it}. This \textit{object-specific reward function} is shared for all objects (\texttt{NN}$^\mathcal{R}$). To evaluate a probability of positive reward $p_i$ for object $i$, it needs to integrate both the corresponding object representation $\mathbf{s}_{obj(i)}$ and the goal. Instead of a simple concatenation, we use a gated-attention mechanism \cite{chaplot2017gatedattention}. $\textbf{g}$ is cast into an attention vector $\balpha^g$ before being combined to $\mathbf{s}_{obj(i)}$ through an Hadamard product (term-by-term): $\mathbf{x}_i^g=\mathbf{s}_{obj(i)} \odot \balpha^g$. The overall architecture is called \textit{MA} for \textit{modular-attention} and can be expressed by:
$$\mathcal{R}(\mathbf{s}, g)~=~\texttt{NN}^\texttt{OR}([\texttt{NN}^\mathcal{R}(\mathbf{s}_{obj(i)} \odot \balpha^g)]_{i\in[1..N]}).$$

   % % % % % % % % % % % % % % % % %
\paragraph{Modular-attention architecture for the policy and critic -}Our agent is controlled by a goal-conditioned policy $\Pi$ that leverages a \textit{modular-attention (MA)} architecture (left in Fig.~\ref{fig:archis}). Similarly, the goal embedding $\mathbf{g}$ is cast into an attention vector $\bbeta^g$ and combined with $\mathbf{s}_{obj(i)}$ through a gated-attention mechanism. As usually done with \deepset, these inputs are projected into a high-dimensional latent space (of size $N \times dim(\mathbf{s}_{obj(i)})$) using a shared network \texttt{NN}$^\Pi$ before being summed. The result is finally mapped into an action vector $\mathbf{a}$ with \texttt{NN}$^\text{a}$. Following the same architecture, the critic (not shown in Fig.~\ref{fig:archis}) computes the action-value of the current state-action pair given the current goal with \texttt{NN}$^\text{av}$: 
\vspace{-0.7cm}
\begin{multicols}{2}
 $$\Pi(\mathbf{s}, g)~=~\texttt{NN}^\text{a}(\sum_{i \in [1..N]}\texttt{NN}^\Pi(\mathbf{s}_{obj(i)} \odot \bbeta^g)),$$
 
  $$Q(\mathbf{s}, \mathbf{a}, g)~=~\texttt{NN}^\text{av}(\sum_{i \in [1..N]}\texttt{NN}^Q([\mathbf{s}_{obj(i)}, \mathbf{a}] \odot \bgamma^g)).$$
\end{multicols}

% % % % % % % % % % % % % % % % % % % % % % % % % % % % % % % % % % % % % % % % % % % % % % % % % % % % % %
\clearpage
\section{Experiments}
% % % % % % % % % % % % % % % % % % % % % % % % % % % % % % % % % % % % % % % % % % % % % % % % % % % % % %

\begin{wrapfigure}[15]{R}{0.6\textwidth}
\vspace{-0.6cm}    
  \subfigure[\label{fig:reward_func}]{\includegraphics[width=0.29\textwidth]{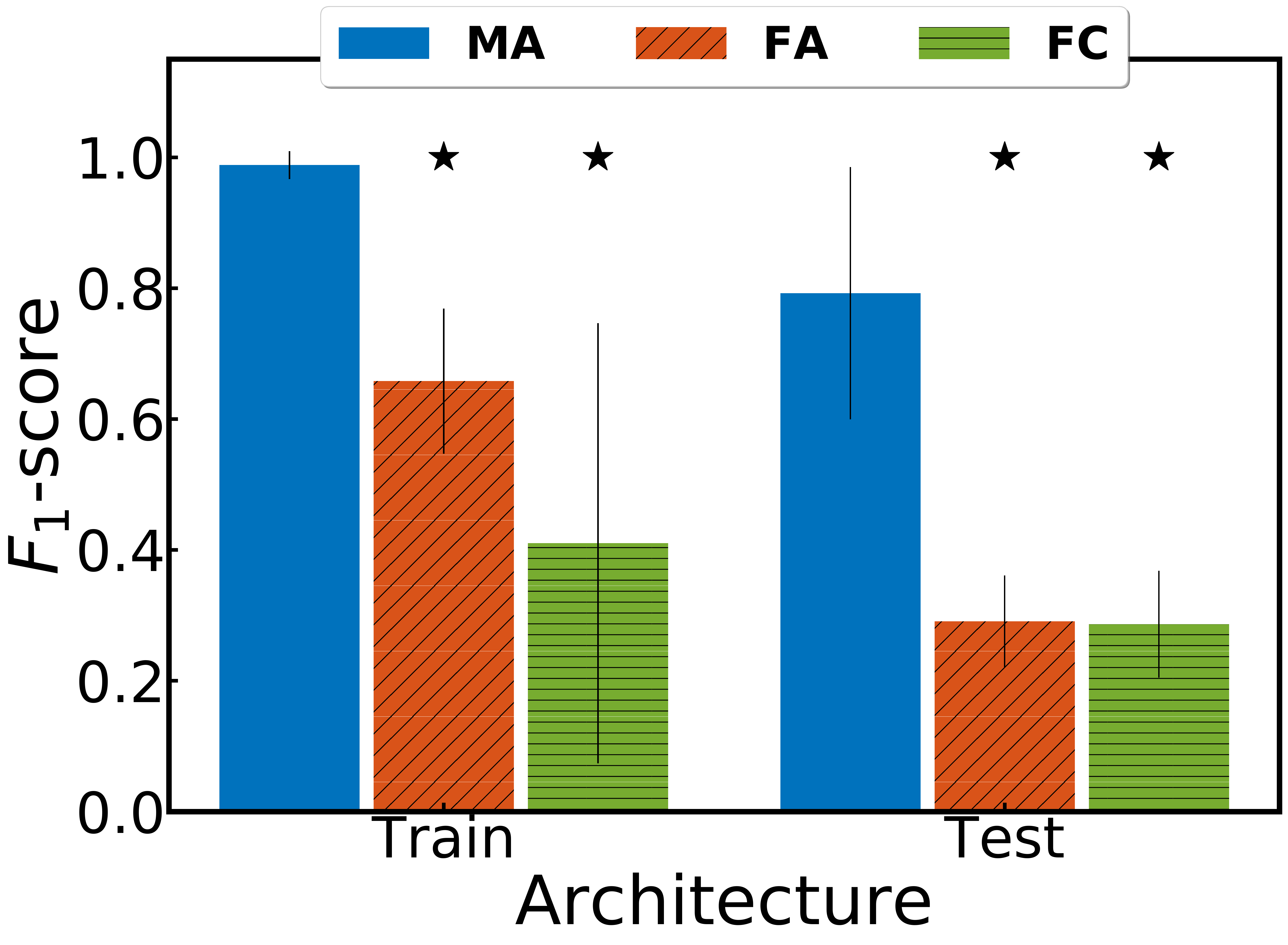}}
      \subfigure[\label{fig:policy_archi}]{\includegraphics[width=0.29\textwidth]{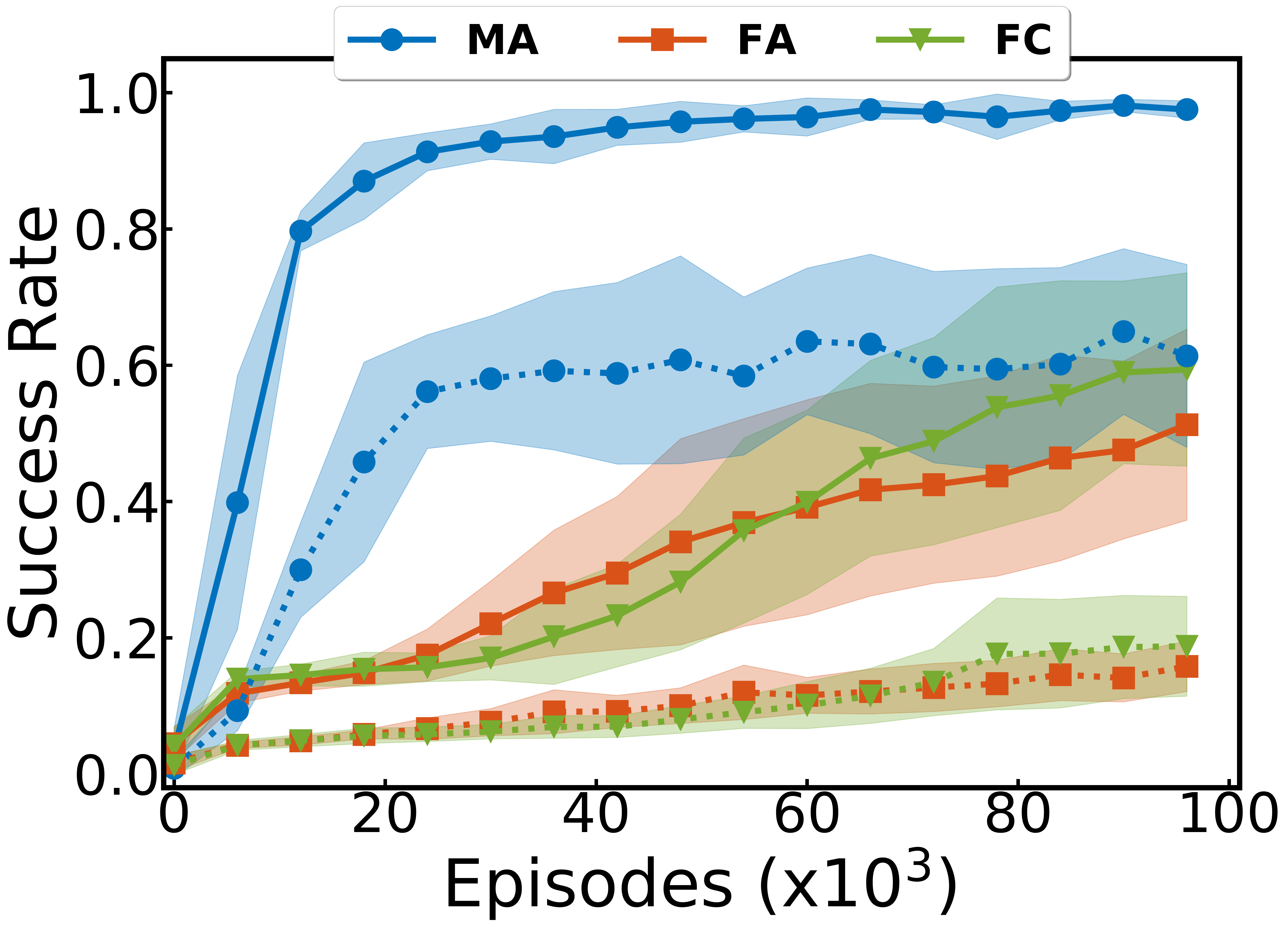}}
      \vspace{-0.4cm}
      \caption{\textbf{Reward function and policy learning.} a: Training (left) and testing (right) performances of the reward function after convergence (stars indicate significant differences w.r.t. \textit{MA}. b: Training (plain) and testing (dashed) performances of the policy. \textit{MA} outperforms \textit{FA} and \textit{FC} on both sets from $ep=600$ $(p<2\cdot10^{-3})$. }
      \label{fig:generalization}
\end{wrapfigure}
\subsection{Generalization Study}
Figure~\ref{fig:generalization} shows the training and testing performances of our proposed \textit{MA} architectures and two baseline architectures: 1) \textit{flat-concatenation (FC)} where the goal embedding is concatenated with the concatenation of object representations and 2) \textit{flat-attention (FA)} where the gated attention mechanism is applied at the scene-level rather than at the object-level (see Fig~\ref{fig:archis} in Appendix). \textit{MA} significantly outperforms competing architectures on both sets. Appendix Section~\ref{sec:reward_per_type} provides detailed generalization performances organized by types of generalizations. 

%\todo{Sample efficiency?}

   % % % % % % % % % % % % % % % % %
\subsection{Robustness to Addition of Objects at Test Time}

\begin{wrapfigure}[11]{R}{0.3\textwidth}
\vspace{-0.5cm}
    \includegraphics[width=0.3\textwidth]{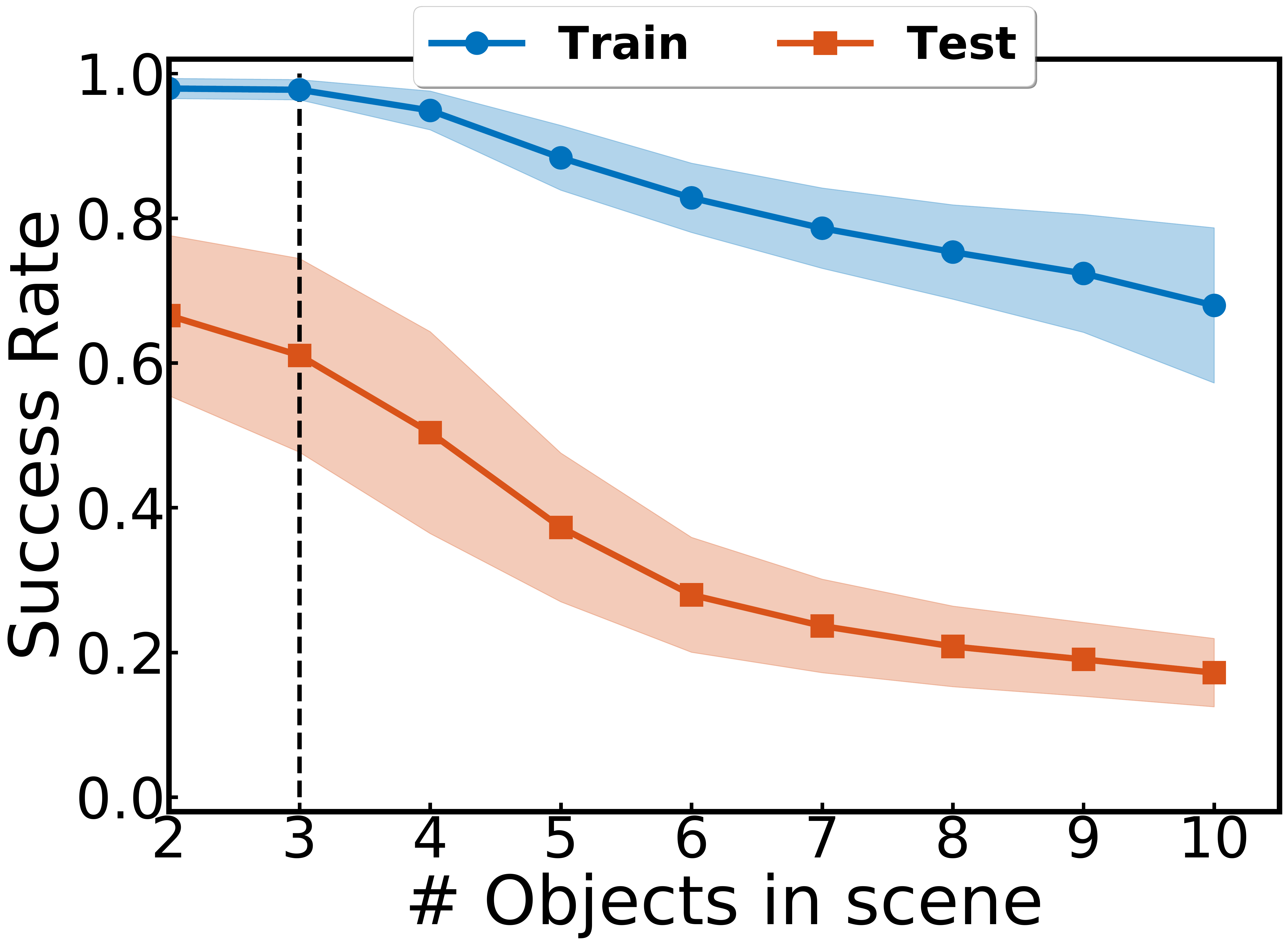}
    \vspace{-0.5cm}
    \caption{\textbf{Varying number of objects at test time.}}
\label{fig:gener_objs}
\end{wrapfigure}
Fully-connected networks using concatenations of object representations are restricted to a constant number of objects $(N)$. In contrast, \textit{MA} architectures treat each object indifferently and in parallel which allows to vary $N$. Whether the performance of the architecture will be affected by $N$ depends on the integration of object-specific information (\textit{OR} for the reward function, \textit{sum} and final network in the policy). Because the OR module is equivalent to a \textit{max} function, it is not affected by $N$ (given a perfect OR). The sum operator merges object-specific information to be used as input of a final network computing the action. As the sum varies with $N$, the overall policy will be sensitive to variations in $N$. Figure~\ref{fig:gener_objs} shows the average training and testing performances of the policy as a function of $N$. We see that a model trained in scenes with $N=3$ objects manages to maintain a reasonable performance on the training set for up to $N=10$, while the generalization performance drops quickly as $N$ increases. $N$ could however be varied during training to make agent robust to its variations.

   % % % % % % % % % % % % % % % % %
\subsection{Introducing Two-Object Interactions}
\label{sec:two_objs}
One can be concerned that the model presented above is limited to object-specific goals. As each module of the reward function receives as input observations from the agent's body and a single object, it cannot integrate multiple-object relationships to estimate the corresponding reward. In this section, we propose to extend the architecture to allow up to two-object relationships. Each module now receives observations from a pair of objects. For $N$ objects, this results in ${N\choose 2}$ modules (e.g. $6$ for $4$ objects). This way, each module is responsible for classifying whether its input pairs verifies the goal or not, while a logical OR integrates this two-object decisions into the global reward.

\begin{wraptable}[7]{R}{0.38\textwidth}
\vspace{-0.4cm}
    \begin{tabular}{c|c|c}
      & 1 obj & 2 objs \\
    \hline
     Train & 0.98 $\pm$ 0.01 & 0.92 $\pm$ 0.02\\
     Test & 0.94 $\pm$ 0.04 & 0.97 $\pm$ 0.02 \\
    \end{tabular}
    \caption{\textbf{$\mathbf{F_1}$-scores on one- and two-object goals.} }
    \label{tab:f1_score_2objs}
\end{wraptable}
To test this, we reuse the dataset described in Section~\ref{sec:supervised-dataset} and relabel its trajectories with one- and two-object goals related to the \textit{grasp} predicate. More specifically, we add goals of the form \textit{grasp} + \textit{any} + \bi{relative position} + \bi{color} $\cup$ \bi{object type} $\cup$ \bi{object category} + \textit{thing}, where \bi{relative position} is one of \{\textit{right\_of, left\_of, above, below}\}. For instance \textit{grasp any right\_of dog thing} is verified whenever the agent grasps an object that was \textit{initially} at the right of any \textit{dog}. These types of goals require to consider two objects: the object to be grasped and the reference object (\textit{dog}). Table~\ref{tab:f1_score_2objs} shows that the reward function can easily be extended to considers object relations. Section~\ref{sec:two_objs_supp} presents a description of the testing set and detailed performances by goal types. % as well as a discussion on relational learning.

   % % % % % % % % % % % % % % % % %
   
% % % % % % % % % % % % % % % % % % % % % % % % % % % % % % % % % % % % % % % % % % % % % % % % % % % % % %
\section{Discussion}
% % % % % % % % % % % % % % % % % % % % % % % % % % % % % % % % % % % % % % % % % % % % % % % % % % % % % %   
In this paper, we investigated how modular architectures of reward function  and policy that operate on unordered sets of object-specific features could benefit generalization. In the context of language-guided autonomous exploration, we showed that the proposed architectures lead to both more efficient learning of behaviors from a training set and improved generalization on a testing set of goals. In addition we investigated generalization to different numbers of objects in the scene at test time and proposed an extension to consider goals related to object pairs.

Humans are known to encode persistent object-specific representations \citep{johnson2013object, green2017object}. Our modular architectures leverage such representations to facilitate transfer of knowledge and skills between object sharing common properties. Although these object features must presently be encoded by engineers, our architectures could be combined with unsupervised multi-object representation learning algorithms \citep{burgess2019monet, greff2019multi}. 

Further work could provide agents the ability to select the number of objects in the scene, from which could emerge a curriculum: if the agent is guided by learning progress, it could first isolate specific objects and their properties, then generalize to more crowded scenes.

%\subsubsection*{Acknowledgments}
%C\'edric Colas and Tristan Karch are partly funded by the French Minist\`ere des Arm\'ees - Direction G\'en\'erale de l’Armement.

\newpage
\bibliography{biblio}

\begin{thebibliography}{22}
\providecommand{\natexlab}[1]{#1}
\providecommand{\url}[1]{\texttt{#1}}
\expandafter\ifx\csname urlstyle\endcsname\relax
  \providecommand{\doi}[1]{doi: #1}\else
  \providecommand{\doi}{doi: \begingroup \urlstyle{rm}\Url}\fi

\bibitem[Andrychowicz et~al.(2017)Andrychowicz, Wolski, Ray, Schneider, Fong,
  Welinder, McGrew, Tobin, Abbeel, and Zaremba]{andrychowicz2017hindsight}
Marcin Andrychowicz, Filip Wolski, Alex Ray, Jonas Schneider, Rachel Fong,
  Peter Welinder, Bob McGrew, Josh Tobin, OpenAI~Pieter Abbeel, and Wojciech
  Zaremba.
\newblock Hindsight experience replay.
\newblock In \emph{Advances in Neural Information Processing Systems}, pp.\
  5048--5058, 2017.

\bibitem[Bahdanau et~al.(2019)Bahdanau, Hill, Leike, Hughes, Kohli, and
  Grefenstette]{bahdanau2018learning}
Dzmitry Bahdanau, Felix Hill, Jan Leike, Edward Hughes, Pushmeet Kohli, and
  Edward Grefenstette.
\newblock {Learning to Understand Goal Specifications by Modelling Reward}.
\newblock In \emph{International Conference on Learning Representations}, jun
  2019.

\bibitem[Battaglia et~al.(2018)Battaglia, Hamrick, Bapst, Sanchez-Gonzalez,
  Zambaldi, Malinowski, Tacchetti, Raposo, Santoro, Faulkner, Gulcehre, Song,
  Ballard, Gilmer, Dahl, Vaswani, Allen, Nash, Langston, Dyer, Heess, Wierstra,
  Kohli, Botvinick, Vinyals, Li, and Pascanu]{battaglia2018relational}
Peter~W. Battaglia, Jessica~B. Hamrick, Victor Bapst, Alvaro Sanchez-Gonzalez,
  Vinicius Zambaldi, Mateusz Malinowski, Andrea Tacchetti, David Raposo, Adam
  Santoro, Ryan Faulkner, Caglar Gulcehre, Francis Song, Andrew Ballard, Justin
  Gilmer, George Dahl, Ashish Vaswani, Kelsey Allen, Charles Nash, Victoria
  Langston, Chris Dyer, Nicolas Heess, Daan Wierstra, Pushmeet Kohli, Matt
  Botvinick, Oriol Vinyals, Yujia Li, and Razvan Pascanu.
\newblock Relational inductive biases, deep learning, and graph networks, 2018.

\bibitem[Bergen(2015)]{bergen2015embodiment}
Benjamin Bergen.
\newblock Embodiment, simulation and meaning.
\newblock \emph{The Routledge handbook of semantics}, pp.\  142--157, 2015.

\bibitem[Burgess et~al.(2019)Burgess, Matthey, Watters, Kabra, Higgins,
  Botvinick, and Lerchner]{burgess2019monet}
Christopher~P Burgess, Loic Matthey, Nicholas Watters, Rishabh Kabra, Irina
  Higgins, Matt Botvinick, and Alexander Lerchner.
\newblock Monet: Unsupervised scene decomposition and representation.
\newblock \emph{arXiv preprint arXiv:1901.11390}, 2019.

\bibitem[Chaplot et~al.(2017)Chaplot, Sathyendra, Pasumarthi, Rajagopal, and
  Salakhutdinov]{chaplot2017gatedattention}
Devendra~Singh Chaplot, Kanthashree~Mysore Sathyendra, Rama~Kumar Pasumarthi,
  Dheeraj Rajagopal, and Ruslan Salakhutdinov.
\newblock Gated-attention architectures for task-oriented language grounding,
  2017.

\bibitem[Colas et~al.(2019{\natexlab{a}})Colas, Oudeyer, Sigaud, Fournier, and
  Chetouani]{curious}
C{\'{e}}dric Colas, Pierre{-}Yves Oudeyer, Olivier Sigaud, Pierre Fournier, and
  Mohamed Chetouani.
\newblock {CURIOUS:} intrinsically motivated modular multi-goal reinforcement
  learning.
\newblock In \emph{Proceedings of the 36th International Conference on Machine
  Learning, {ICML} 2019, 9-15 June 2019, Long Beach, California, {USA}}, pp.\
  1331--1340, 2019{\natexlab{a}}.

\bibitem[Colas et~al.(2019{\natexlab{b}})Colas, Sigaud, and
  Oudeyer]{colas2019hitchhiker}
C{\'e}dric Colas, Olivier Sigaud, and Pierre-Yves Oudeyer.
\newblock A hitchhiker's guide to statistical comparisons of reinforcement
  learning algorithms.
\newblock \emph{arXiv preprint arXiv:1904.06979}, 2019{\natexlab{b}}.

\bibitem[Colas et~al.(2020)Colas, Karch, Lair, Dussoux, Moulin-Frier, Dominey,
  and Oudeyer]{imagine}
Cédric Colas, Tristan Karch, Nicolas Lair, Jean-Michel Dussoux, Clément
  Moulin-Frier, Peter~Ford Dominey, and Pierre-Yves Oudeyer.
\newblock Language as a cognitive tool to imagine goals in curiosity-driven
  exploration.
\newblock 2020.

\bibitem[Fu et~al.(2019)Fu, Korattikara, Levine, and Guadarrama]{fu2018from}
Justin Fu, Anoop Korattikara, Sergey Levine, and Sergio Guadarrama.
\newblock {From Language to Goals: Inverse Reinforcement Learning for
  Vision-Based Instruction Following}.
\newblock In \emph{International Conference on Learning Representations}, 2019.

\bibitem[Green \& Quilty-Dunn(2017)Green and Quilty-Dunn]{green2017object}
Edwin~James Green and Jake Quilty-Dunn.
\newblock What is an object file?
\newblock \emph{The British Journal for the Philosophy of Science}, 2017.

\bibitem[Greff et~al.(2019)Greff, Kaufmann, Kabra, Watters, Burgess, Zoran,
  Matthey, Botvinick, and Lerchner]{greff2019multi}
Klaus Greff, Rapha{\"e}l~Lopez Kaufmann, Rishab Kabra, Nick Watters, Chris
  Burgess, Daniel Zoran, Loic Matthey, Matthew Botvinick, and Alexander
  Lerchner.
\newblock Multi-object representation learning with iterative variational
  inference.
\newblock \emph{arXiv preprint arXiv:1903.00450}, 2019.

\bibitem[Hochreiter \& Schmidhuber(1997)Hochreiter and
  Schmidhuber]{hochreiter1997lstm}
Sepp Hochreiter and J\"{u}rgen Schmidhuber.
\newblock Long short-term memory.
\newblock \emph{Neural Comput.}, 9\penalty0 (8):\penalty0 1735–1780, November
  1997.
\newblock ISSN 0899-7667.
\newblock \doi{10.1162/neco.1997.9.8.1735}.
\newblock URL \url{https://doi.org/10.1162/neco.1997.9.8.1735}.

\bibitem[Johnson et~al.(2017)Johnson, Hariharan, Maaten, Fei-Fei, Zitnick, and
  Girshick]{Johnson_2017}
Justin Johnson, Bharath Hariharan, Laurens van~der Maaten, Li~Fei-Fei,
  C.~Lawrence Zitnick, and Ross Girshick.
\newblock Clevr: A diagnostic dataset for compositional language and elementary
  visual reasoning.
\newblock \emph{2017 IEEE Conference on Computer Vision and Pattern Recognition
  (CVPR)}, Jul 2017.
\newblock \doi{10.1109/cvpr.2017.215}.
\newblock URL \url{http://dx.doi.org/10.1109/CVPR.2017.215}.

\bibitem[Johnson(2013)]{johnson2013object}
Scott~P Johnson.
\newblock Object perception.
\newblock \emph{Handbook of developmental psychology}, pp.\  371--379, 2013.

\bibitem[Kaschak \& Glenberg(2000)Kaschak and
  Glenberg]{kaschak2000constructing}
Michael~P Kaschak and Arthur~M Glenberg.
\newblock Constructing meaning: The role of affordances and grammatical
  constructions in sentence comprehension.
\newblock \emph{Journal of memory and language}, 43\penalty0 (3):\penalty0
  508--529, 2000.

\bibitem[Mankowitz et~al.(2018)Mankowitz, {\v{Z}}{\'\i}dek, Barreto, Horgan,
  Hessel, Quan, Oh, van Hasselt, Silver, and Schaul]{mankowitz2018unicorn}
Daniel~J Mankowitz, Augustin {\v{Z}}{\'\i}dek, Andr{\'e} Barreto, Dan Horgan,
  Matteo Hessel, John Quan, Junhyuk Oh, Hado van Hasselt, David Silver, and Tom
  Schaul.
\newblock Unicorn: Continual learning with a universal, off-policy agent.
\newblock \emph{arXiv preprint arXiv:1802.08294}, 2018.

\bibitem[Santoro et~al.(2017)Santoro, Raposo, Barrett, Malinowski, Pascanu,
  Battaglia, and Lillicrap]{santoroRBMPBL17}
Adam Santoro, David Raposo, David G.~T. Barrett, Mateusz Malinowski, Razvan
  Pascanu, Peter~W. Battaglia, and Timothy~P. Lillicrap.
\newblock A simple neural network module for relational reasoning.
\newblock \emph{CoRR}, abs/1706.01427, 2017.
\newblock URL \url{http://arxiv.org/abs/1706.01427}.

\bibitem[Vaswani et~al.(2017)Vaswani, Shazeer, Parmar, Uszkoreit, Jones, Gomez,
  Kaiser, and Polosukhin]{vaswani2017attention}
Ashish Vaswani, Noam Shazeer, Niki Parmar, Jakob Uszkoreit, Llion Jones,
  Aidan~N Gomez, {\L}ukasz Kaiser, and Illia Polosukhin.
\newblock Attention is all you need.
\newblock In \emph{Advances in neural information processing systems}, pp.\
  5998--6008, 2017.

\bibitem[Wagstaff et~al.(2019)Wagstaff, Fuchs, Engelcke, Posner, and
  Osborne]{wagstaff2019limitations}
Edward Wagstaff, Fabian~B Fuchs, Martin Engelcke, Ingmar Posner, and Michael
  Osborne.
\newblock On the limitations of representing functions on sets.
\newblock \emph{arXiv preprint arXiv:1901.09006}, 2019.

\bibitem[Zaheer et~al.(2017)Zaheer, Kottur, Ravanbakhsh, Poczos, Salakhutdinov,
  and Smola]{deepset}
Manzil Zaheer, Satwik Kottur, Siamak Ravanbakhsh, Barnabas Poczos, Russ~R
  Salakhutdinov, and Alexander~J Smola.
\newblock Deep sets.
\newblock In \emph{Advances in neural information processing systems}, pp.\
  3391--3401, 2017.

\bibitem[Zambaldi et~al.(2018)Zambaldi, Raposo, Santoro, Bapst, Li, Babuschkin,
  Tuyls, Reichert, Lillicrap, Lockhart, Shanahan, Langston, Pascanu, Botvinick,
  Vinyals, and Battaglia]{zambaldi2018relational}
Vinicius Zambaldi, David Raposo, Adam Santoro, Victor Bapst, Yujia Li, Igor
  Babuschkin, Karl Tuyls, David Reichert, Timothy Lillicrap, Edward Lockhart,
  Murray Shanahan, Victoria Langston, Razvan Pascanu, Matthew Botvinick, Oriol
  Vinyals, and Peter Battaglia.
\newblock Relational deep reinforcement learning, 2018.

\end{thebibliography}
\bibliographystyle{iclr2020_conference}

\newpage
\appendix
\section{Environment and Grammar}
\subsection{Environment}

\begin{wrapfigure}[17]{R}{0.45\textwidth}
% \vspace{0.4cm}
    \includegraphics[width=.45\textwidth]{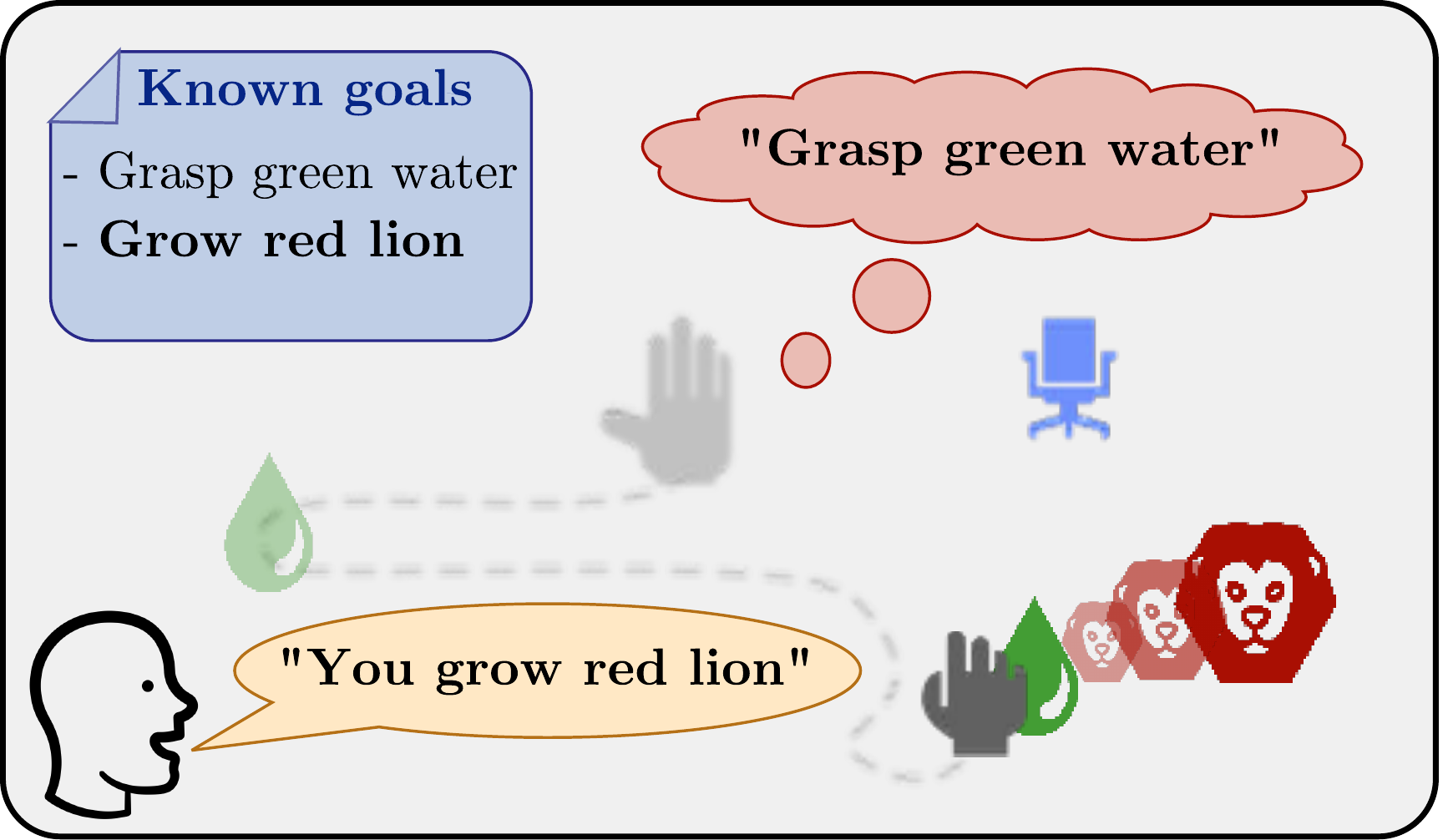}
        \caption{\textbf{The \textit{Playground} environment.} The agent targets a goal represented as NL and receives descriptive feedback from the SP to expand its repertoire of known goals}
\label{fig:env_descr}
\end{wrapfigure}

\begin{wrapfigure}[13]{R}{0.45\textwidth}
% \vspace{0.4cm}
    \includegraphics[width=.45\textwidth]{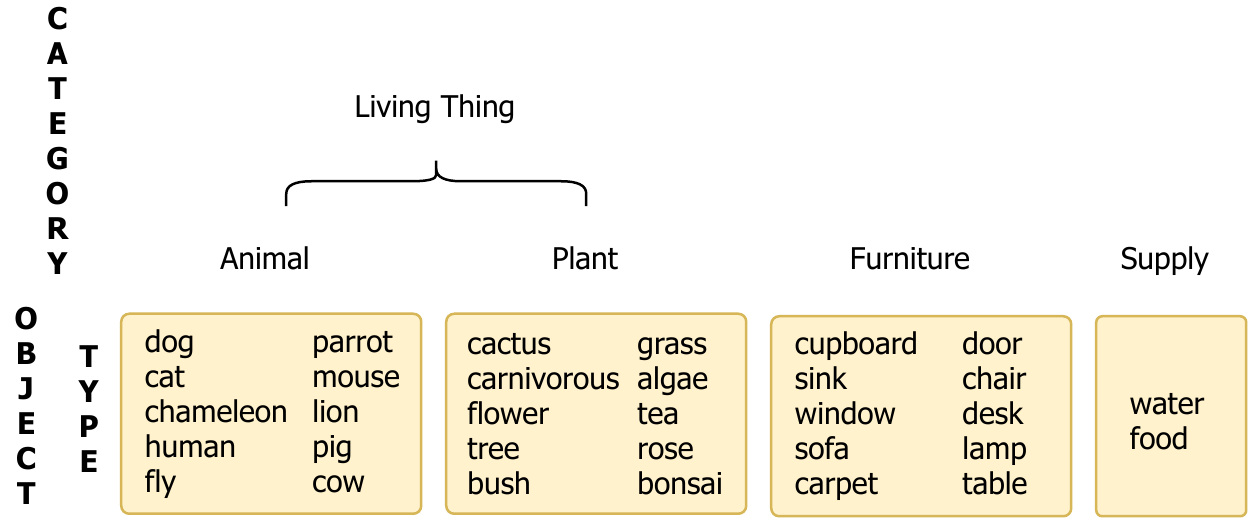}
        \caption{\textbf{Representation of possible objects types and categories.}}
\label{fig:env_category}
\end{wrapfigure}

The \textit{Playground} environment is a continuous $2$D world. In each episode, $N=3$ objects are uniformly sampled from a set of $32$ different object types (\textit{e.g. dog, cactus, sofa, water, etc.}), organized into $5$ categories (\textit{animals, furniture, plants, etc.}), see Fig.~\ref{fig:env_category}. Sampled objects have a color (R,G,B) and can be grasped. Animals and plants can be grown when the right supplies are brought to them (food or water for animal, water for plants), whereas furniture cannot (e.g. sofa). Random scene generations are conditioned by the goals selected by the agent (e.g. \textit{grasp red lion} requires the presence of a \textit{red lion}). 

\textit{Agent embodiment - }In this environment, the agent can perform bounded continuous translations in the $2$D plane, grasp and release objects by changing the state of its gripper. It perceives the world from an allocentric perspective and thus has access to the whole scene. 

\textit{Agent perception - }The scene is described by a state vector containing information about the agent's body and the $N$ objects. Each object is represented by a set of features describing its type (one-hot encoding of size $32$), its $2$D-position, color (\textsc{RGB} code), size (scalar) and whether it is grasped (boolean). Categories are not explicitly encoded. Color, size and initial positions are sampled from uniform distributions making each object unique. At time step $t$, we can define an observation $\textbf{o}_t$ as the concatenation of body observations ($2$D-position, gripper state) and objects' features. The state $\textbf{s}_t$ used as input of the models is the concatenation of $\textbf{o}_t$ and $\bDelta\textbf{o}_t = \textbf{o}_t-\textbf{o}_0$.

\textit{Social partner - }Part of the environment, SP is implemented by a hard-coded function taking the final state of an episode ($\mathbf{s}_T$) as input and returning NL descriptions of $\mathbf{s}_T$: $\mathcal{D}_{\text{SP}}(\mathbf{s}_T) \subset \mathcal{D}^\text{SP}$. When SP provides \textit{descriptions}, the agent hears targetable \textit{goals}. Given the set of previously discovered goals $(\mathcal{G}_d)$ and new descriptions $\mathcal{D}_{\text{SP}}(\mathbf{s}_T)$, the agent infers the set of goals that were not achieved: $\mathcal{G}_{na}(\mathbf{s}_T)~=~\mathcal{G}_{d}~\backslash~\mathcal{D}_{SP}(\mathbf{s}_T)$, where $\backslash$ indicates complement.

\subsection{Grammar}
\label{sec:grammar}
\begin{multicols}{2}
\begin{enumerate}[leftmargin=0.6cm, noitemsep]
    \item Go: \textit{(e.g. go bottom left)} %\vspace{-0.2cm}
    \begin{itemize} [leftmargin=0.2cm, noitemsep]
            \item \textit{go} + \bi{zone}
            \end{itemize}
    \item Grasp: \textit{(e.g. grasp red cat)} %\vspace{-0.2cm}
        \begin{itemize}[leftmargin=0.2cm, noitemsep]
            \item \textit{grasp} + \bi{color} $\cup$ \{\textit{any}\}  + \textbf{\textit{object type $\cup$ object category}}
            \item \textit{grasp} + \textit{any} + \bi{color} + \textit{thing}
        \end{itemize}
    \item Grow: \textit{(e.g. grow green animal)} %\vspace{-0.2cm}
        \begin{itemize}[leftmargin=0.2cm, noitemsep]
            \item \textit{grow} + \bi{color} $\cup$ \{\textit{any}\} + \bi{living thing} $\cup$ \{\textit{living\_thing, animal, plant}\}
            \item \textit{grow} + \textit{any} + \bi{color} + \textit{thing}
    \end{itemize}
\end{enumerate} 
\end{multicols}
\bi{zone} includes words referring to areas of the scene (e.g. \textit{top}, \textit{right}, \textit{bottom left}), \bi{object type} is one of $32$ object types (e.g. \textit{parrot}, \textit{cactus}) and \bi{object category} one of $5$ object categories (\textit{living\_thing, animal, plant, furniture, supply}). \bi{living thing} refers to any plant or animal word, \bi{color} is one of \textit{blue, green, red} and \textit{any} refers to any color, or any object. 

\newpage
\begin{wraptable}[41]{R}{0.5\textwidth}
\begin{tabular}{c|l}
        \multirow{2}{2.75em}{Type 1} & \textit{Grasp blue door}, \textit{Grasp green dog}, \\ 
                                        & \textit{Grasp red tree}, \textit{Grow green dog} \\ 
        \hline
        \multirow{4}{2.75em}{Type 2} & \textit{Grasp any flower}, \textit{Grasp blue flower},\\
                                     & \textit{Grasp green flower}, \textit{Grasp red flower},\\
                                     & \textit{Grow any flower}, \textit{Grow blue flower},\\
                                     & \textit{Grow green flower}, \textit{Grow red flower},\\
        \hline
        \multirow{2}{2.75em}{Type 3} & \textit{Grasp any animal}, \textit{Grasp blue animal}, \\  
                                     & \textit{Grasp green animal}, \textit{Grasp red animal} \\ 
        \hline
        \multirow{2}{2.75em}{Type 4} & \textit{Grasp any fly}, \textit{Grasp blue fly}, \\ 
                                    & \textit{Grasp green fly}, \textit{Grasp red fly} \\ 
        \hline
        \multirow{22}{2.75em}{Type 5} & \textit{Grow any algae}, \textit{Grow any bonsai}\\
                                     & \textit{Grow any bush}, \textit{Grow any cactus}\\
                                     & \textit{Grow any carnivorous}, \textit{Grow any grass}\\
                                     & \textit{Grow any living\_thing}, \textit{Grow any plant}\\
                                     & \textit{Grow any rose}, \textit{Grow any tea}\\
                                     & \textit{Grow any tree}, \textit{Grow blue algae}\\
                                     & \textit{Grow blue bonsai}, \textit{Grow blue bush}\\
                                     & \textit{Grow blue cactus}, \textit{Grow blue carnivorous}\\
                                     & \textit{Grow blue grass}, \textit{Grow blue living\_thing}\\
                                     & \textit{Grow blue plant}, \textit{Grow blue rose}\\
                                     & \textit{Grow blue tea}, \textit{Grow blue tree}\\
                                     & \textit{Grow green algae}, \textit{Grow green bonsai}\\
                                     & \textit{Grow green bush}, \textit{Grow green cactus}\\
                                     & \textit{Grow green carnivorous}, \textit{Grow green grass}\\
                                     & \textit{Grow green living\_thing}, \textit{Grow green plant}\\
                                     & \textit{Grow green rose}, \textit{Grow green tea}\\
                                     & \textit{Grow green tree}, \textit{Grow red algae}\\
                                     & \textit{Grow red bonsai}, \textit{Grow red bush}\\
                                     & \textit{Grow red cactus}, \textit{Grow red carnivorous}\\
                                     & \textit{Grow red grass}, \textit{Grow red living\_thing}\\
                                     & \textit{Grow red plant}, \textit{Grow red rose}\\
                                     & \textit{Grow red tea}, \textit{Grow red tree}\\
    \end{tabular}
    \caption{Testing goals in $\mathcal{G}^\text{test}$}
    \label{tab:test_descriptions}
\end{wraptable}
\section{5 Types of Generalization}
We define $5$ different types of \textit{out-of-distribution} generalization:
%\vspace{-0.4cm}
\begin{itemize}[leftmargin=0.6cm,noitemsep]
    \item Type 1 - \textit{Attribute-object generalization}: \bi{predicate} + \textit{\{blue door, red tree, green dog\}}. Understanding \textit{grasp red tree} requires to leverage knowledge about the \textit{red} attribute (grasping red non-tree objects) and the \textit{tree} object type (grasping non-red tree objects). 
    \item Type 2 - \textit{Attribute extrapolation}:  \bi{predicate} + \bi{color}  $\cup$ \{\textit{any}\} + \textit{flower}. As \textit{flower} is removed from the training set, \textit{grasp red flower} requires the extrapolation of the red attribute to a new object type.
    \item Type 3 - \textit{Predicate-category generalization}: \textit{grasp} + \bi{color} $\cup$ \{\textit{any}\} + \textit{animal}. Understanding \textit{grasp any animal} requires to understand the animal category (from growing \textit{animal} and growing animal objects) and the \textit{grasp} predicate (from grasping non-\textit{animal} objects) to transfer the former to the latter.
    \item Type 4 - \textit{Easy predicate-object generalization}:  \textit{grasp} + \bi{color} $\cup$ \{\textit{any}\} + \{\textit{fly}\}. Understanding \textit{grasp any fly} requires to leverage knowledge about the \textit{grasp} predicate (grasping non-fly objects) and the \textit{fly} object (growing flies).
    \item Type 5 - \textit{Hard predicate-object generalization}: \textit{grow} + \bi{color} $\cup$ \{\textit{any}\} + \bi{plant}  $\cup$ \{\textit{plant, living\_thing}\}. \textit{grow any plant} requires to understand the \textit{grow} predicate (from growing animals) and the \bi{plant} objects (and category) (from grasping plant objects). However, this transfer is more complex than the reverse transfer in Type 4 for two reasons. First, the interaction modalities vary: plants only grow with \textit{water}. Second, Type 4 is only about the \textit{fly} object, while here it is about all \bi{plant} objects and the \textit{plant} and \textit{living\_thing} categories.
\end{itemize}
Each of the testing goals described above is removed from the training set ($\mathcal{G}^\text{train} \cap \mathcal{G}^\text{test} = \emptyset$). Table~\ref{tab:test_descriptions} provides the complete list of testing goals. 
\newpage
\section{Dataset}
\label{sec:supervised-dataset}
\begin{figure}[!h]
    \centering
        \includegraphics[width=0.4\textwidth]{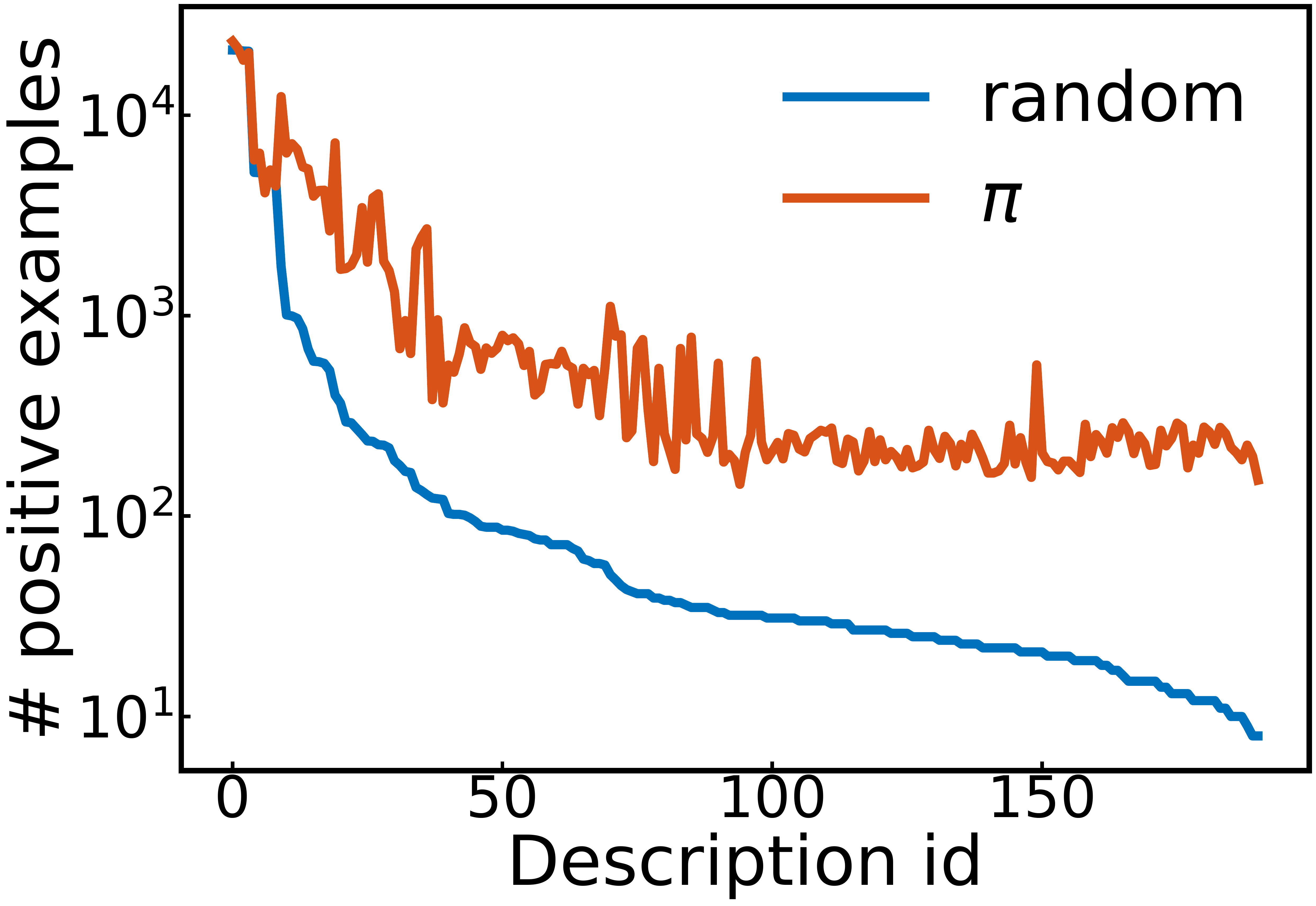}
         \caption{\textbf{Data distributions for the supervised learning of the reward function.} Sorted counts of positive examples per training set descriptions.}  
    \label{fig:count_reward}
\end{figure}

The reward function is trained in two contexts. First in a supervised setting, independently from the policy. Second, it is trained in parallel of the policy during RL runs. To learn a reward function in a supervised setting, we first collected a dataset of $50\times 10^3$ trajectories and associated goal descriptions using a random action policy. Training the reward function on this data led to poor performances, as the number of positive examples remained low for some goals (see Fig.~\ref{fig:count_reward}). To pursue the independent analysis of the reward function, we used $50\times 10^3$ trajectories collected by an RL agent co-trained with its reward function using \textit{modular-attention} architectures (data characterized by the top distribution in Fig.~\ref{fig:count_reward}). Results presented in Fig.~\ref{fig:reward_func} used such RL-collected data. To closely match the training conditions imposed by the co-learning setting, we train the reward function on the final states $s_T$ of each episode and test it on any state $s_t$ for $t=[1,...,T]$ of other episodes. The performance of the reward function are crucial to jointly learn the policy.

\clearpage
\newpage
\section{Architecture}
\label{sec:architecture}

\begin{figure}[!ht]
\centering
    \includegraphics[width=.7\columnwidth]{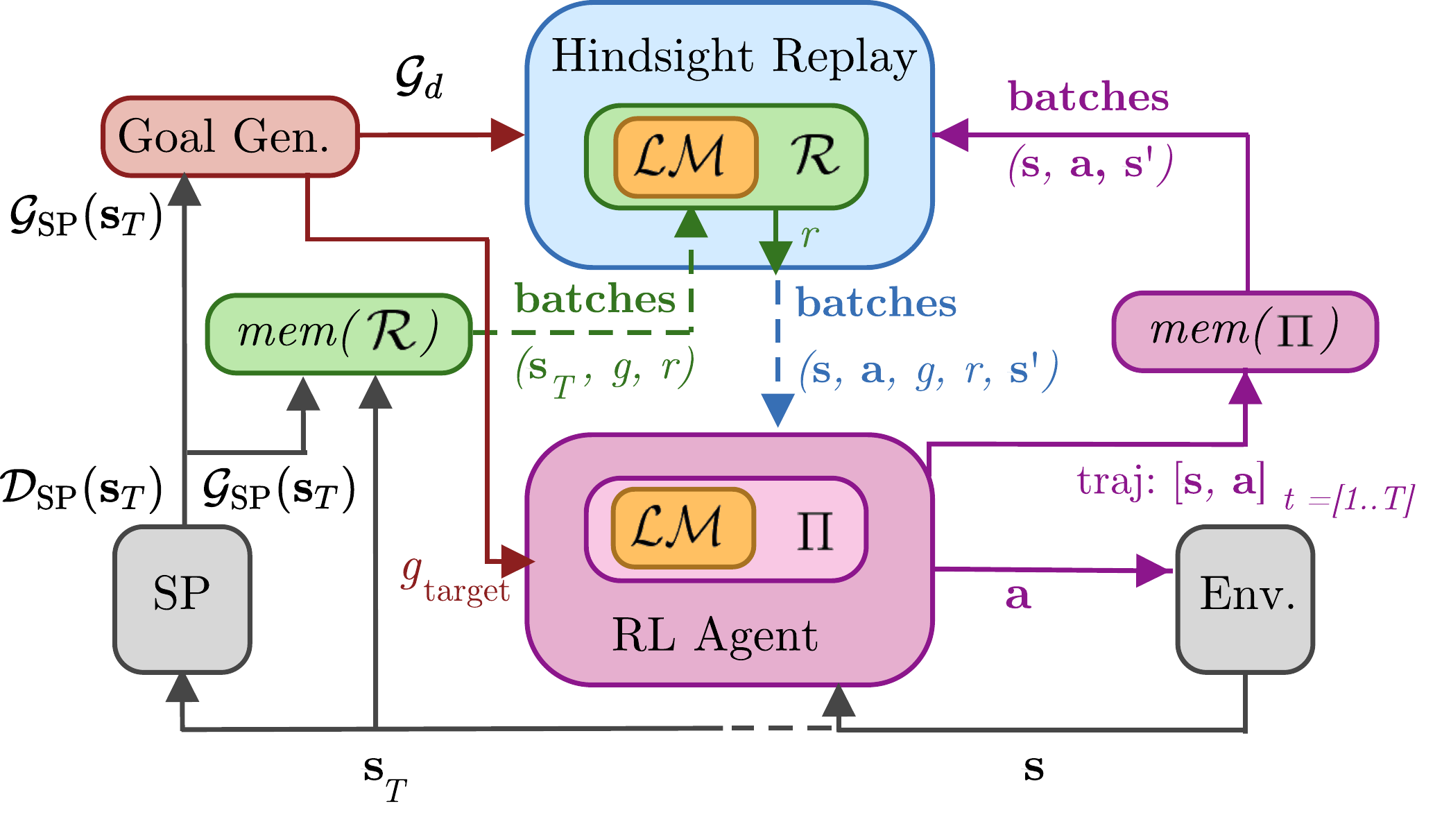}
     \caption{\textbf{The \imagine architecture.} Colored boxes represent the different modules composing \imagine. Lines represent update signals (dashed) and function outputs (plain). $\mathcal{LM}$ is shared.}  
\label{fig:architecture}
\end{figure}

Figure~\ref{fig:architecture} represents the \imagine architecture whose logic can be outlined as follows:  
%\vspace{-0.2cm}
\begin{enumerate}[leftmargin=0.6cm, nolistsep]
    \item The \textit{Goal Generator} samples a target goal $g_{\text{target}}$ from discovered goals $(\mathcal{G}_d)$.
    \item The agent interacts with the environment (\textit{RL Agent}) using its policy $\Pi$ conditioned by $g_\text{target}$.
    \item The state-action trajectories are stored in \textit{mem$(\Pi)$}.
    \item SP observes $\mathbf{s}_T$ and provides descriptions $\mathcal{D}_{\text{SP}}(\textbf{s}_T)$ that the agent turns into targetable goals $\mathcal{G}_{\text{SP}}(\textbf{s}_T)$.
    \item \textit{mem$(\mathcal{R})$} stores positive pairs $(\mathbf{s}_T, \mathcal{G}_{\text{SP}}(\mathbf{s}_T))$ and infers negative pairs $(\mathbf{s}_T, \mathcal{G}_{\text{na}}(\mathbf{s}_T))$.
    \item The agent then updates:
    %\vspace{-0.1cm}
    \begin{itemize}[leftmargin=0.2cm,noitemsep]
        \item \textit{Goal Generator}: $\mathcal{G}_{d}~\gets~\mathcal{G}_{d} \cup \mathcal{G}_{\text{SP}}(\textbf{s}_T)$.
        \item \textit{Language Model} $(\mathcal{LM})$ and \textit{Reward Function} $(\mathcal{R})$: updated using data from \textit{mem$(\mathcal{R})$}.
        \item \textit{RL agent} (actor and critic): A batch of state-action transitions $(\mathbf{s},\mathbf{a},\mathbf{s}')$ is sampled from \textit{mem$(\Pi)$}. Then \textit{Hindsight Replay} and $\mathcal{R}$ are used to select goals to train on and compute rewards $(\mathbf{s},\mathbf{a},\mathbf{s},g_{\text{NL}},r)$. Finally, the policy and critic are trained via RL.
    \end{itemize}
\end{enumerate}
Descriptions of the language model, reward function and policy can be found in the main article. Next paragraphs describe others modules. Further implementation details, training schedules and pseudo-code can be found in the companion paper \citep{imagine}.

\paragraph{Language model -}The language model $(\mathcal{LM})$ embeds NL goals $(\mathcal{LM}(g_\text{NL}):\mathcal{G}^\text{NL} \to \mathbb{R}^{100})$ using an \texttt{LSTM} \citep{hochreiter1997lstm} trained jointly with the reward function (yellow in Fig.~\ref{fig:archis}). The reward function, policy and critic become language-conditioned functions when coupled with $\mathcal{LM}$, which acts like a goal translator.

\paragraph{Modular Reward Function using Deep Sets -} Learning a goal-conditioned reward function $(\mathcal{R})$ is framed as a binary classification. The reward function maps a state $\mathbf{s}$ and a goal embedding $\mathbf{g}=\mathcal{LM}(g_\text{NL})$ to a binary reward: $\mathcal{R}(\mathbf{s}, \mathbf{g}): \mathcal{S} \times \mathbb{R}^{100} \to \{ 0, 1\}$ (left in Fig.~\ref{fig:archis}). 

\textit{Architecture - } The reward function, policy and critic leverage modular architectures inspired by \deepset \citep{deepset} combined with gated attention mechanisms \citep{chaplot2017gatedattention}. \deepset is a network architecture implementing set functions (input permutation invariance). Each input is mapped separately to some (usually high-dimensional \citep{wagstaff2019limitations}) latent space using a shared network. These latent representations are then passed through a permutation-invariant function (e.g. mean, sum) to ensure the permutation-invariance of the whole function. 
In the case of our reward function, inputs are grouped into object-dependent sub-states $\mathbf{s}_{obj(i)}$, each mapped to a probability $p_i$ by a same network \texttt{NN}$^\mathcal{R}$ (weight sharing). \texttt{NN}$^\mathcal{R}$ can be thought of as a single-object reward function which estimates whether object $i$ verifies the goal $(p_i>0.5)$ or not. Probabilities $[p_i]_{i\in[1..N]}$ for the $N$ objects are then mapped into a global binary reward using a logical \texttt{OR} function: \textit{if any object verifies the goal, then the whole scene verifies it}. This \texttt{OR} function implements object-wise permutation-invariance. In addition to object-dependent inputs, the computation of $p_i$ integrates goal information through a gated-attention mechanism. Instead of being concatenated, the goal embedding $\textbf{g}$ is cast into an attention vector $\balpha^g$ before being combined to the object-dependent sub-state through an Hadamard product (term-by-term) to form the inputs of \texttt{NN}$^\mathcal{R}$: $\mathbf{x}_i^g=\mathbf{s}_{obj(i)} \odot \balpha^g$. This can be seen as scaling object-specific features according to the interpretation of the goal $g_\text{NL}$. Finally, we pre-trained a neural-network-based \texttt{OR} function: \texttt{NN}$^\texttt{OR}$ such that the output is $1$ whenever $\text{max}_i([p_i]_{i\in[1..N]}) > 0.5$. This is required to enable end-to-end training of $\mathcal{LM}$ and $\mathcal{R}$. The overall function can be expressed by:
$$\mathcal{R}(\mathbf{s}, g)~=~\texttt{NN}^\texttt{OR}([\texttt{NN}^\mathcal{R}(\mathbf{s}_{obj(i)} \odot \balpha^g)]_{i\in[1..N]}).$$
We call this architecture \textit{MA} for \textit{modular-attention}. 

\textit{Data - } Interacting with the environment and SP, the agent builds a dataset of triplets $(\mathbf{s}_T, \mathbf{g}, r)$ where $r$ is a binary reward marking the achievement of $\mathbf{g}$ in state $\mathbf{s}_T$. $\mathcal{LM}$ and $\mathcal{R}$ are periodically updated by backpropagation on this dataset.

   % % % % % % % % % % % % % % % % %
\paragraph{Modular Policy using Deep Sets -}Our agent is controlled by a goal-conditioned policy $\Pi$ that leverages an adapted \textit{modular-attention (MA)} architecture (right in Fig.~\ref{fig:archis}). Similarly, the goal embedding $\mathbf{g}$ is cast into an attention vector $\bbeta^g$ and combined with the object-dependent sub-state $\mathbf{s}_{obj(i)}$ through a gated-attention mechanism. As usually done with \deepset, these inputs are projected into a high-dimensional latent space (of size $N \times dim(\mathbf{s}_{obj(i)})$) using a shared network \texttt{NN}$^\Pi$ before being summed. The result is finally mapped into an action vector $\mathbf{a}$ with \texttt{NN}$^\text{a}$. Following the same architecture, the critic computes the action-value of the current state-action pair given the current goal with \texttt{NN}$^\text{av}$: 
\vspace{-0.7cm}
\begin{multicols}{2}
 $$\Pi(\mathbf{s}, g)~=~\texttt{NN}^\text{a}(\sum_{i \in [1..N]}\texttt{NN}^\Pi(\mathbf{s}_{obj(i)} \odot \bbeta^g)),$$
 
  $$Q(\mathbf{s}, \mathbf{a}, g)~=~\texttt{NN}^\text{av}(\sum_{i \in [1..N]}\texttt{NN}^Q([\mathbf{s}_{obj(i)}, \mathbf{a}] \odot \bgamma^g)).$$
\end{multicols}

% % % % % % % % % % % % 

\paragraph{Hindsight learning -}Our agent uses \textit{hindsight learning}, which means it can \textit{replay} the memory of a trajectory (e.g. when trying to grasp object A) by \textit{pretending} it was targeting a different goal (e.g. grasping object B) \citep{andrychowicz2017hindsight, mankowitz2018unicorn, curious}. In practice, goals originally targeted during data collection are replaced by others in the batch of transitions used for RL updates, a technique known as \textit{hindsight replay} \citep{andrychowicz2017hindsight}. To generate candidate substitute goals, we use the reward function to scan a list of $50$ goals sampled randomly so as to bias the ratio of positive examples.

% % % % % % % % % % % % 

\paragraph{Goal generator -}Generated goals are used to serve as targets during environment interactions and as substitute goals for hindsight replay. The goal generator samples uniformly from the set of discovered goals $\mathcal{G}_d$.

\newpage
\section{Competing Architectures}

\begin{figure*}[!hb]
      \centering
       \subfigure[\label{fig:concat}]{\includegraphics[width=0.49\textwidth]{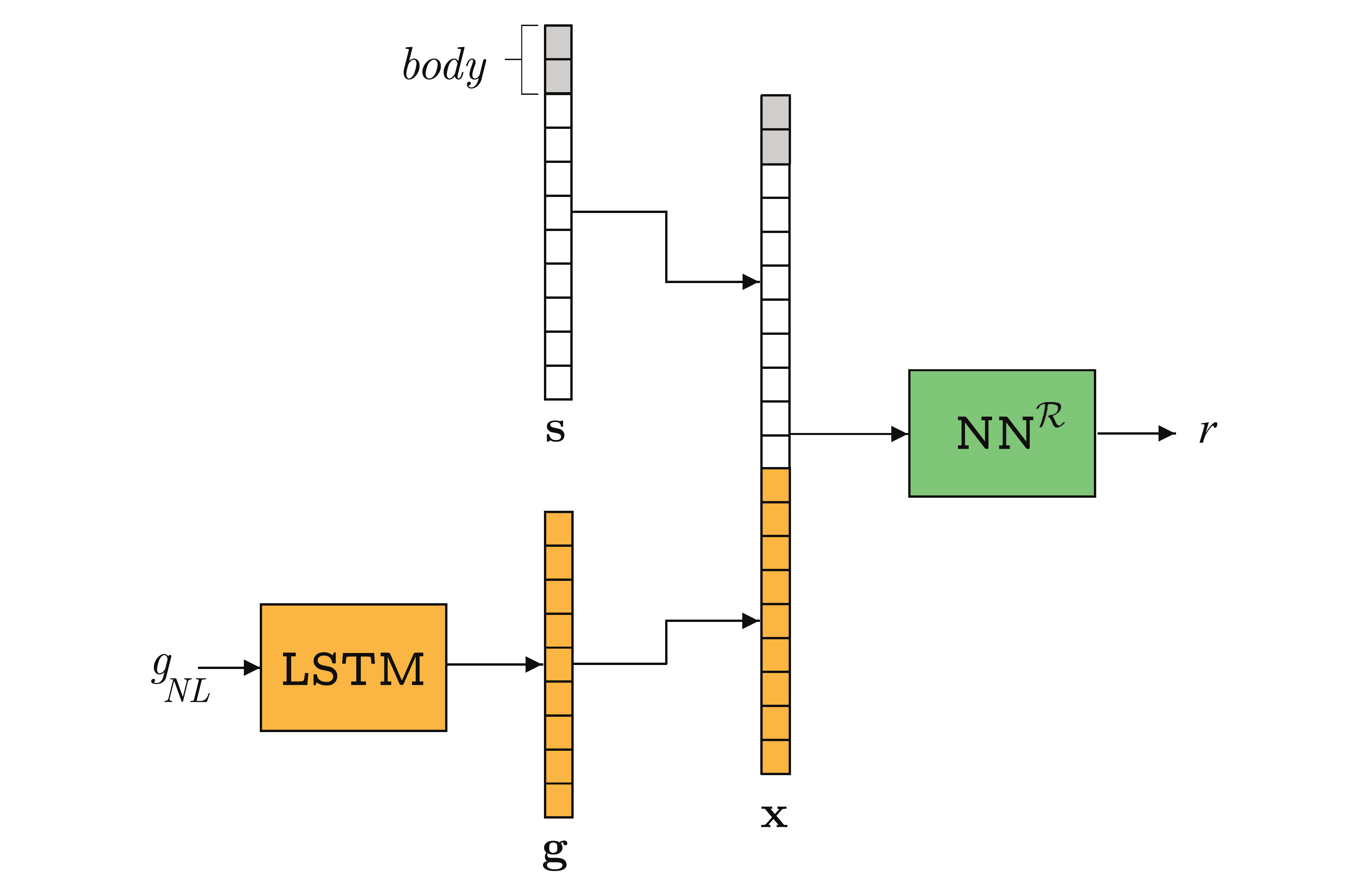}}
      \subfigure[\label{fig:attention}]{\includegraphics[width=0.49\textwidth]{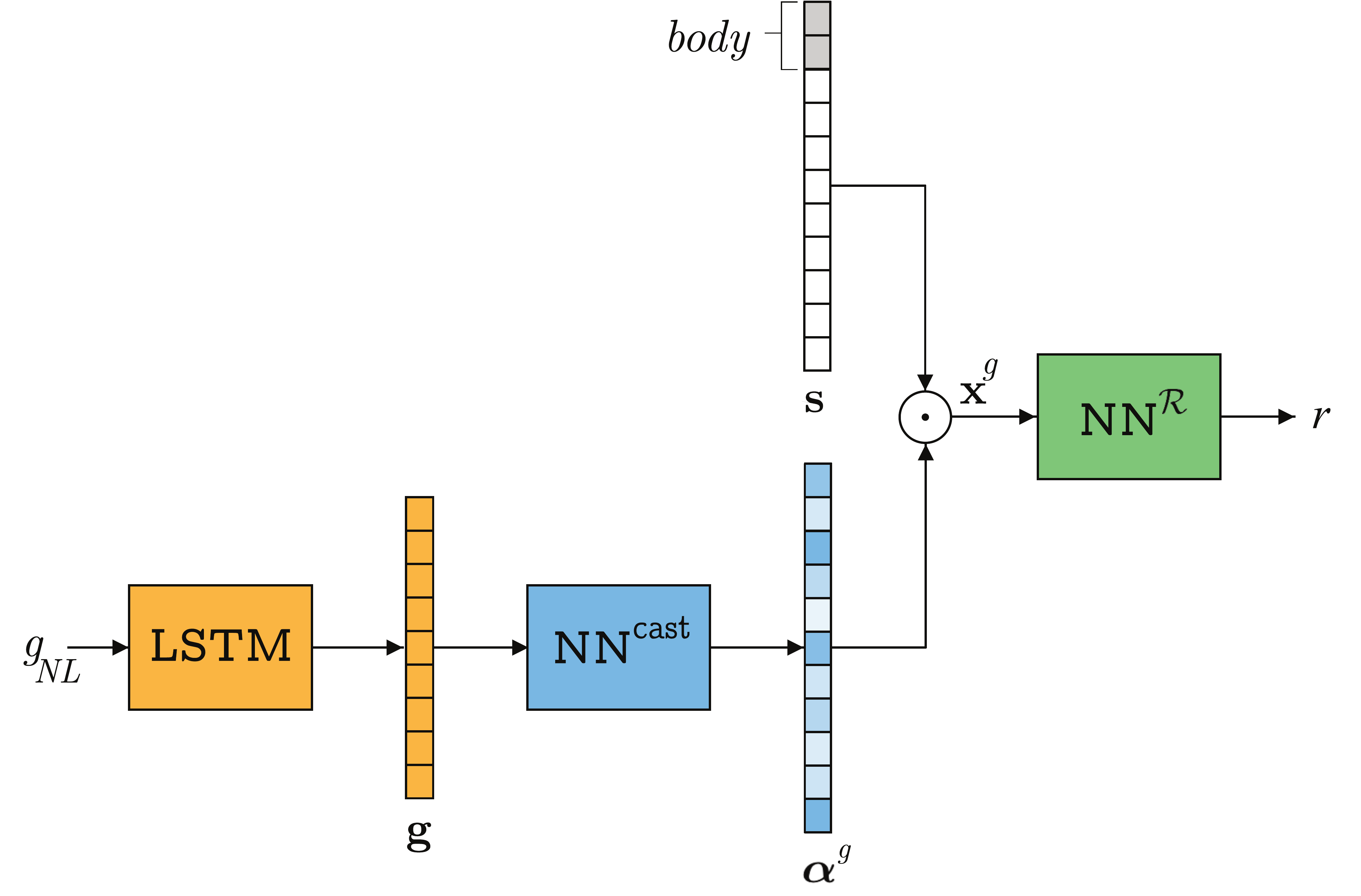}}
      \caption{\textbf{Competing architectures.} a: Flat-concatenation (FC). b: Flat-attention (FA).}
      \label{fig:competing_archi}
\end{figure*} 

\newpage
\section{Results: Generalization per Type}
\label{sec:reward_per_type}
\begin{figure}[h!]
    \centering
    \subfigure[\label{fig:pol-gene-per-type}]{\includegraphics[width=0.45\textwidth]{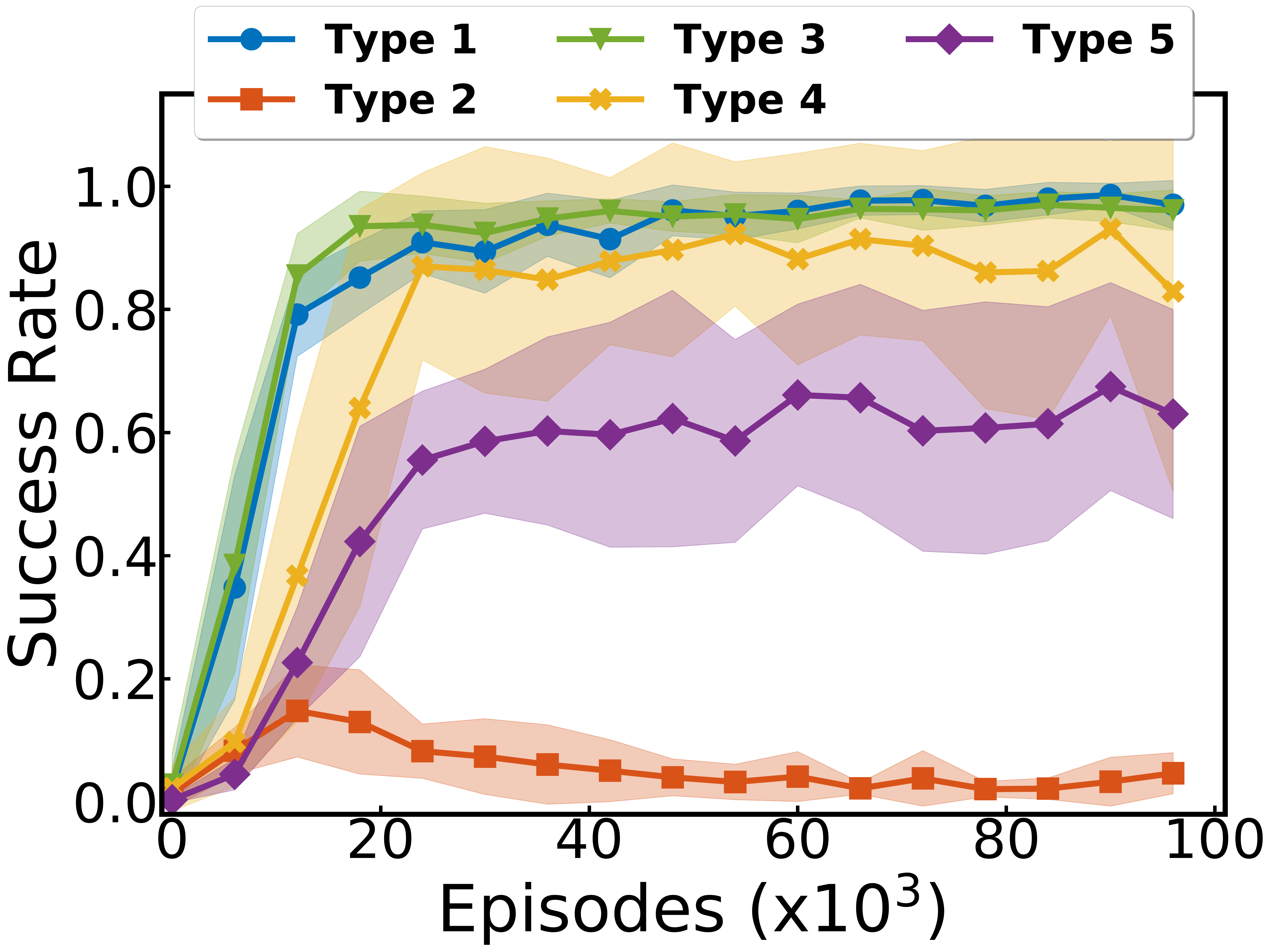}}
    \subfigure[\label{fig:rf-gene-per-type}]{\includegraphics[width=0.45\textwidth]{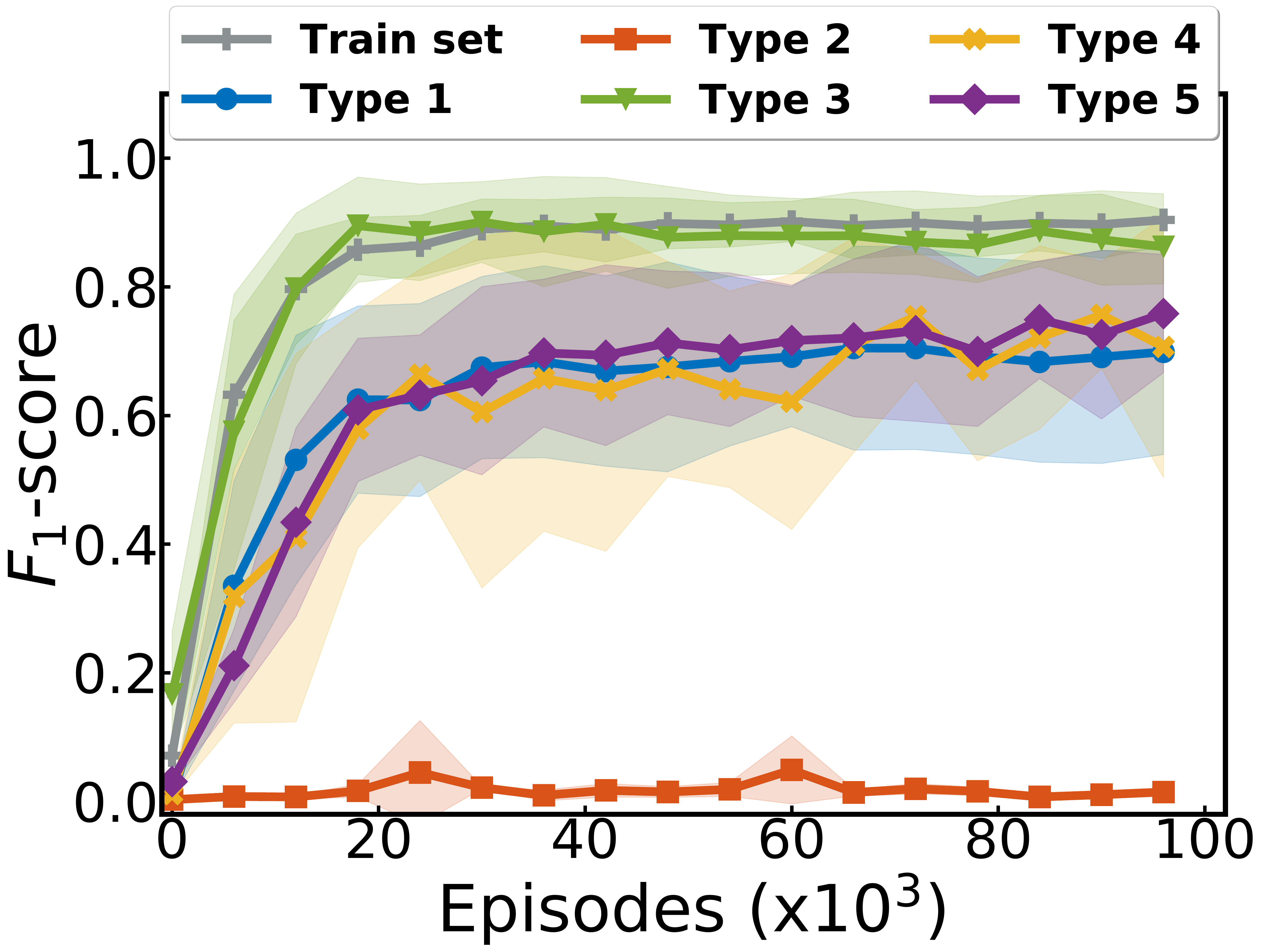}}
    \caption{\textbf{Policy and reward function generalization.} a: Average success rate of the policy. b: $F_1$ score of the reward function.}
    \label{fig:gene-per-type}
\end{figure}

Fig.~\ref{fig:pol-gene-per-type} provides the average success rate for the five generalization types. \textit{MA} models demonstrate good generalizations of Type 1 (\textit{attribute-object generalization}, e.g. \textit{grasp red tree}), Type 3 (\textit{predicate-category generalization}, e.g. \textit{grasp any animal}) and Type 4 (\textit{easy predicate-object generalization}: e.g. \textit{grasp any fly}). Generalizing the meaning \textit{grow} to other objects (Type 5, \textit{hard predicate-object generalization}) is harder as it requires to understand the dynamics of the environment. As we could expect, the generalization of colors to new objects fails (Type 2, \textit{attribute extrapolation}). As Type 2 introduces a new word, the language model's \texttt{LSTM} receives a new token, which perturbs the encoding of the sentence. The generalization capabilities of the reward function when it is jointly trained with the policy are provided in Fig.~\ref{fig:rf-gene-per-type}. They seem to be inferior to the policy's capabilities, especially for Type 1 and 4. It should however be noted that the $F_1$-score plotted in Fig.~\ref{fig:rf-gene-per-type} does not necessarily describe the actual generalization that occurs during the joint training of the reward function and the policy as it is computed from the supervised learning trajectories (see Section~\ref{sec:supervised-dataset}).

\newpage
\section{Two-Object Results}
\label{sec:two_objs_supp}

\begin{figure}[ht]
    \begin{minipage}{0.45\linewidth}
    \centering
    \includegraphics[width=\textwidth]{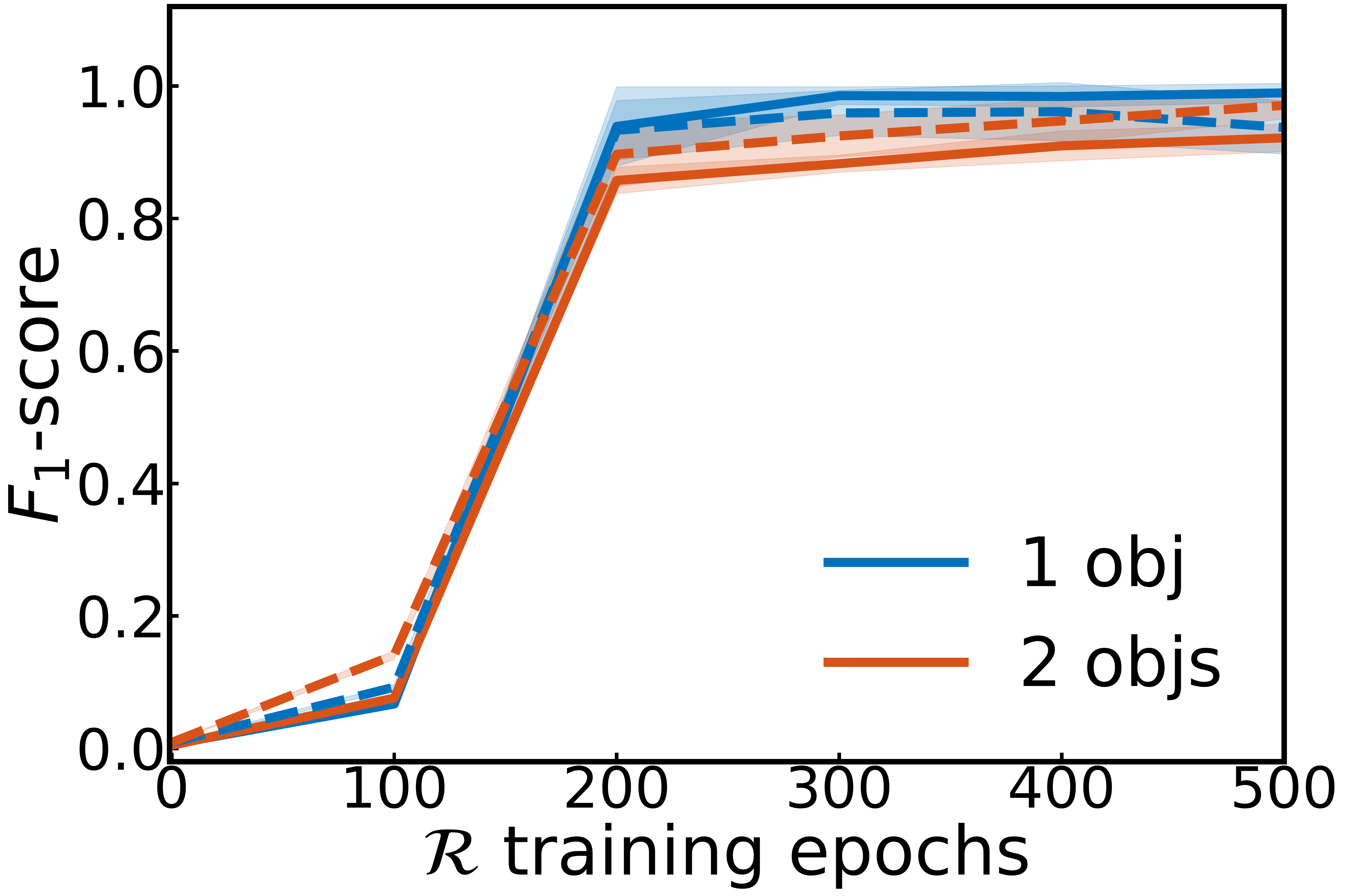}
    \caption{Convergence plot of the reward function. $F_1$-score w.r.t training epochs computed over the training (plain) and testing (dashed) sets for one-object goals (blue) and two-objects goals (red).}
    \label{fig:2objs_res}
    \end{minipage}
\hspace{0.5cm}
    \begin{minipage}{0.45\linewidth}
    \vspace{-1.2cm}
    \centering
    \begin{tabular}{c|l}
        \multirow{4}{2.75em}{1 obj} & \textit{Grasp any animal}, \\
        & \textit{Grasp blue animal}, \\
        & \textit{Grasp red animal}, \\
        &\textit{Grasp green animal} , \\
        & \textit{Grasp any fly}, \textit{Grasp blue fly} \\ 
        & \textit{Grasp red fly}, \textit{Grasp green fly}, \\
        & \textit{Grasp blue door}, \\
        & \textit{Grasp green dog}, \\
        & \textit{Grasp red tree} \\
        \hline
        \multirow{2}{2.75em}{2 objs} & \textit{Grasp any left\_of blue thing},\\
                                     & \textit{Grasp any right\_of dog thing}\\
    
    \end{tabular}
    \vspace{0.5cm}
    \caption{Test goals used for the object-pair analysis.}
    \label{tab:test_goals_2objs}
    \end{minipage}
\end{figure}

Fig.~\ref{fig:2objs_res} shows the evolution of the $F_1$-score of the reward function computed from the training set and the testing set (given in Fig.~\ref{tab:test_goals_2objs}). The model considering two-objects interactions exhibit near perfect $F_1$-score for both one-object goals and two-objects goals. Note that, after convergence, the testing $F_1$-score is higher than the training one for two-objects goals. This is due to the fact that the testing set for two-objects goals is limited to only two examples.

\vspace{0.5cm}

%\paragraph{A word on relation learning -}In both the reward function and the policy, the final decision (whether binary reward or action vector) integrates sub-decisions taken at the object-level. Every object-level decision takes into account relationships between the body --seen as a very special kind of object-- and either one or a pair of external objects. This is the heart of language grounding, understanding language sentences in terms of one's body (controllable through actions) and external objects (controllable via the body). 

\end{document}